\documentclass[runningheads]{llncs}
\usepackage{graphicx}
\usepackage[dvipsnames]{xcolor}

\begin{document}
\title{Interactive Classification for\\Deep Learning Interpretation}
%
%

\author{\'{A}ngel Cabrera \and
Fred Hohman \and
Jason Lin \and
Duen Horng Chau}
\authorrunning{Cabrera et al.}
%
\institute{Georgia Institute of Technology, Atlanta GA, USA\\
\email{\{alex.cabrera, fredhohman, jlin401, polo\}@gatech.edu}
}
\maketitle              
\begin{abstract}
We present an interactive system enabling users to manipulate images to explore the robustness and sensitivity of deep learning image classifiers. 
Using modern web technologies to run in-browser inference, users can remove image features using inpainting algorithms and obtain new classifications in real time,
which allows them to ask a variety of ``what if'' questions by experimentally modifying images and seeing how the model reacts.
Our system allows users to compare and contrast what image regions humans and machine learning models use for classification,
revealing a wide range of surprising results ranging from spectacular failures (e.g., a \textit{water bottle} image becomes a \textit{concert} when removing a person) to impressive resilience (e.g., a \textit{baseball player} image remains correctly classified even without a glove or base).
We demonstrate our system at The 2018 Conference on Computer Vision and Pattern Recognition (CVPR) for the audience to try it live.
Our system is open-sourced at \url{https://github.com/poloclub/interactive-classification}.
A video demo is available at \url{https://youtu.be/llub5GcOF6w}. 

\end{abstract}
\section{Introduction}

Public trust in artificial intelligence and machine learning is essential to its prevalence and widespread acceptance.
To create trust, both researchers and the general public have to understand why models behave the way they do.
Existing research has used interactive data visualization as a mechanism for humans to interface with black-box machine learning models~\cite{hohman2018visual}, revealing how models learn and behave using an interactive dialogue~\cite{weld2018intelligible} as opposed to static explanations.

\subsection{Interactive Classification for Interpretation}
We have designed and developed an interactive system that allows users to experiment with deep learning image classifiers and explore their robustness and sensitivity.
Users are able to remove selected areas of an image in real time with classical computer vision inpainting algorithms, Telea~\cite{telea2004image} and PatchMatch~\cite{barnes2009patchmatch}, which allows them to ask a variety of ``what if'' questions by experimentally modifying images and seeing how the model reacts~\cite{hohman2017shapeshop}.
Allowing a human user to select the regions to be inpainted enables more semantically meaningful regions to be considered for study compared to other techniques where shapes (e.g., gray squares) are placed over random parts of an image to subdue signal to the image classifier~\cite{zeiler2014visualizing}.
Some existing work uses automated image segmentation techniques, e.g., superpixels~\cite{achanta2012slic}, and a combinatorial search algorithm to find the most important superpixels; however, there is no guarantee the superpixels are semantically meaningful features to consider~\cite{ribeiro2016should}.

\subsection{Significance of Our Approach}
Through interactive inpainting, our system helps reveal a wide range of surprising results ranging from spectacular failures (e.g., a \textit{water bottle} image becomes a \textit{concert} when removing a person) to impressive resilience (e.g., a \textit{baseball player} image remains correctly classified even without a glove or base).
The system also computes class activation maps on demand, which highlight the important semantic regions of an image a model uses for classification~\cite{zhou2015cnnlocalization}.
Combining these tools, users can develop qualitative insight into what a model sees and which features impact an image's classification.
Our system is open-sourced at \url{https://github.com/poloclub/interactive-classification}.
A video demo is available at \url{https://youtu.be/llub5GcOF6w}.

\begin{figure}[t]
\begin{center}
\includegraphics[width=1\textwidth]{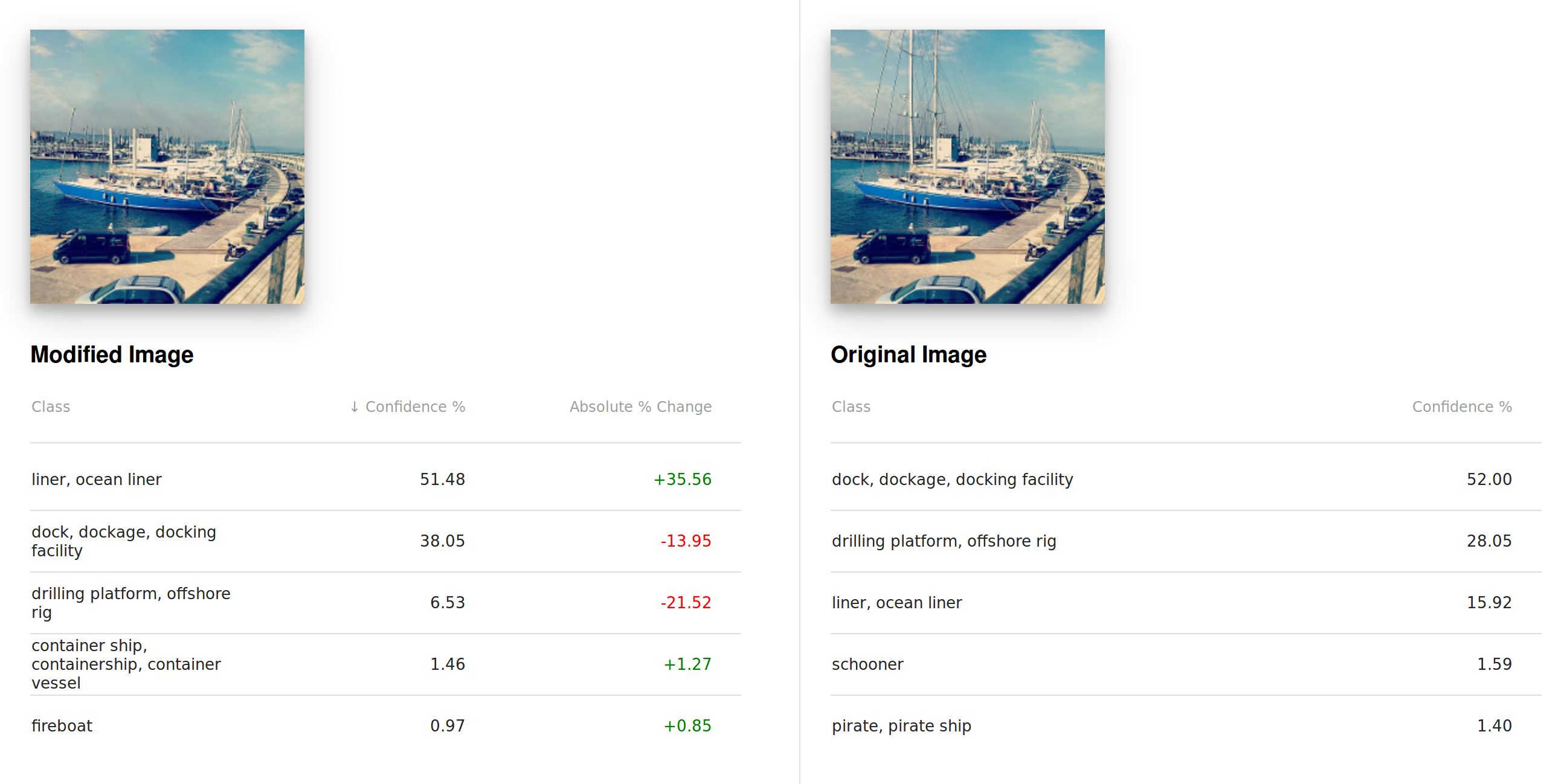}
\end{center}
   \caption{
   The modified image (left), originally classified as \textit{dock} is misclassified as \textit{ocean liner} when the masts of a couple boats are removed from the original image (right).
   The top five classification scores are tabulated underneath each image.
   }
\label{fig:boat}
\end{figure}

\section{Usage Examples}

\subsection{Misclassification}
In Figure \ref{fig:boat} we present an example of our interactive classification system with a deep learning image classifier that exhibits failures when given semantic edits.
Consider an image of a couple cars driving next to boats moored at a dock.
The original top 5 classification results for this image are shown on the right in Figure \ref{fig:boat}.
The classifier originally correctly produces a top class of \textit{dock}.
Using our system, the user highlights the masts of the boats, inpainting them and replacing them with sky. 
The edits do not alter the meaning of the picture, but unfortunately the classification changes to an inaccurate class.
With the masts removed, the modified image (Figure \ref{fig:boat} left) is now misclassified as an \textit{ocean liner} instead of the correct classification of a \textit{dock} (Figure \ref{fig:boat} right).

\begin{figure}[tbh]
\begin{center}
\includegraphics[width=1\textwidth]{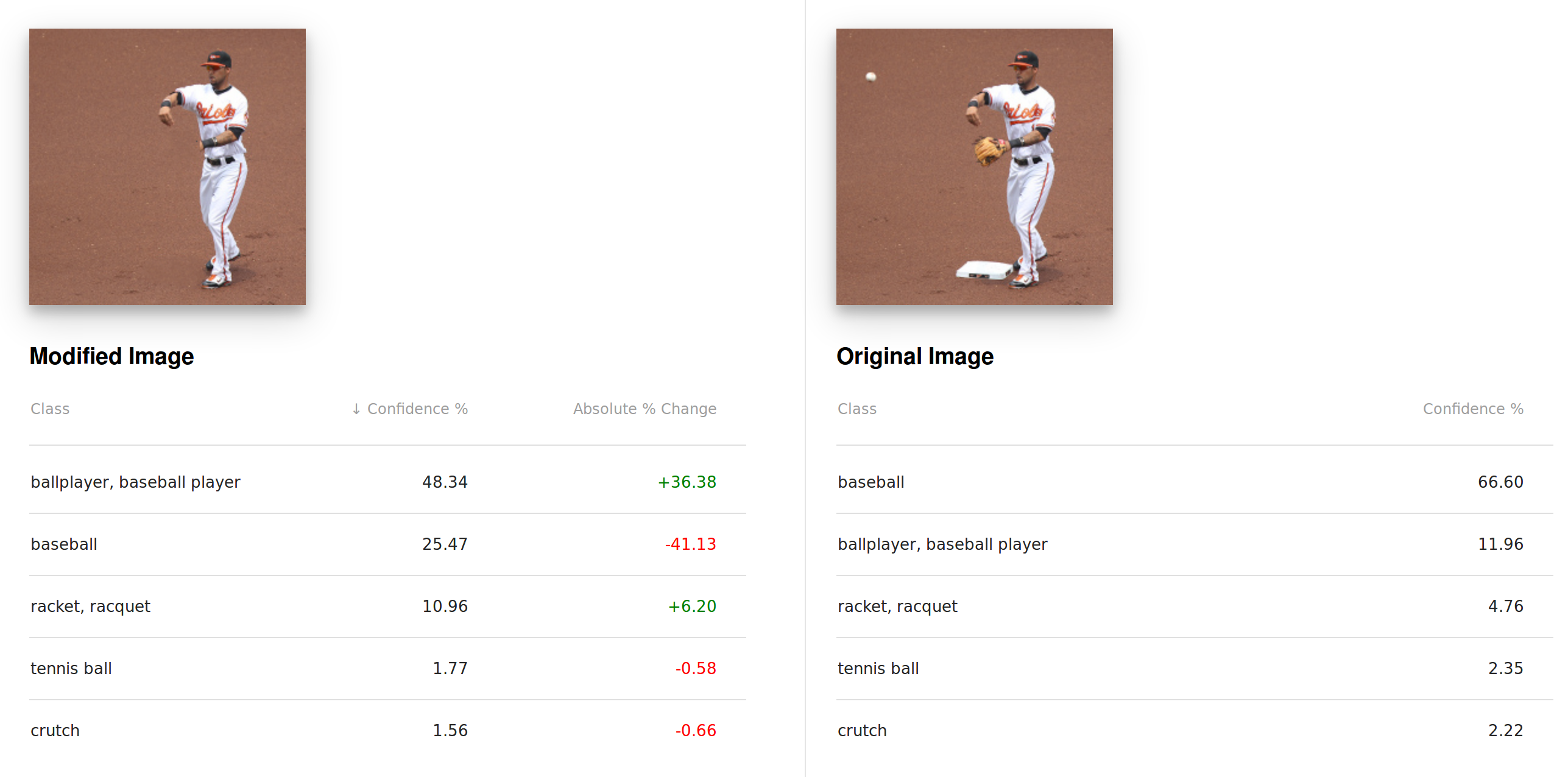}
\end{center}
   \caption{
    The modified image (left), originally classified as \textit{baseball} and \textit{ballplayer} remains correctly classified when the ball, glove, and base are removed from the original image (right).
    The top five classification scores are tabulated underneath each image with decreasing confidence.
   }
\label{fig:baseball}
\end{figure}

\subsection{Resilient Classification}
In Figure \ref{fig:baseball} we present an example of our interactive classification system with a deep learning image classifier that exhibits resilience to semantic edits.
Consider an image instance of a baseball player with a glove throwing a ball at base.
The original top 5 classification results for this image are shown on the right in Figure \ref{fig:baseball}.
The top 2 class are \textit{baseball} and \textit{ballplayer}.
Using our system, the user highlights the baseball, glove, and base, inpainting and replacing them with the brown dirt.
Surprisingly, the modified image remains classified as \textit{baseball player} when these important features are removed from the original image (Figure \ref{fig:baseball} right), albeit with adjusted probabilities (Figure \ref{fig:baseball} left).

\section{System Design}
To provide an easily accessible experience, we created an in-browser system by combining multiple novel web technologies.
React.js forms the base of the system, providing an interactive and approachable interface that encourages image manipulation. 
The other core technology is Tensorflow.js, which runs deep learning models in the browser and can classify images in near real-time.
We used the SqueezeNet~\cite{iandola2016squeezenet} and MobileNet~\cite{howard2017mobilenets} deep learning models for Tensorflow.js since they are designed to be portable and fast.
Lastly, we make two different inpainting algorithms available to the user: Telea~\cite{telea2004image} and PatchMatch~\cite{barnes2009patchmatch}.
The Telea algorithm uses a fast, heuristic-based method and is implemented fully in JavaScript, allowing the system to be run completely in-browser.
For better, more advanced inpainting we utilize the GIMP Resynthesizer plugin, which is an open-source implementation of the PatchMatch algorithm.
This is implemented in C and exposed to the system through a Python server.

\section{Conclusion}
We present an interactive system that enables users to explore the robustness and sensitivity of deep learning image classifiers. 
Using modern web technologies to run in-browser inference and compute class activation maps, users can inpaint images in real time, to ask a variety of ``what if'' questions by experimentally modifying images and seeing how the model reacts.
Our system helps reveal a wide range of surprising classification results ranging from spectacular failures to impressive resilience.
Our system is open-sourced. 
We believe our investigation will help people explore the extent to which humans and machines think alike, and shed light on the advantages and potential pitfalls of deep learning image classifiers.

\section*{Acknowledgments}
This work was supported by NSF grants IIS-1563816, CNS-1704701, and TWC-1526254; NASA Space Technology Research Fellowship; and gifts from Google, Symantec, Yahoo, Intel, Microsoft, eBay, Amazon.
%
%
%
\bibliographystyle{splncs04}
\bibliography{main}
\end{document}